\documentclass[9pt,twocolumn,twoside]{pnas-new}
\templatetype{pnasresearcharticle} % Choose template
\begin{document}

\title{Two-cluster test}

% Use letters for affiliations, numbers to show equal authorship (if applicable) and to indicate the corresponding author
\author[a]{Xinying Liu}
\author[a]{Lianyu Hu}
\author[a]{Mudi Jiang}
\author[a]{Simeng Zhang}
\author[a]{Jun Lou}
\author[a,1]{Zengyou He}

\affil[a]{School of Software, Dalian University of Technology, Dalian, China.}

\leadauthor{Xingying Liu}

% Please add a significance statement to explain the relevance of your work
\significancestatement{A crucial issue in many clustering methods is to determine whether two subsets of samples belong to the same cluster. However, existing two-sample tests are not appropriate for solving this issue since subsets are not given in advance but generated by clustering algorithms. Here we introduce the two-cluster test, which incorporates the fact that two subsets are homogenous groups provided by some clustering procedures. By choosing a proper test statistic, we can derive an analytical $p$-value to quantify whether ``two clusters should be separated or not". Some empirical studies demonstrate that it is a must to investigate such a new statistical test and it can be deployed as the key component in different types of clustering issues.}

% Please include corresponding author, author contribution and author declaration information
\authorcontributions{X.L. designed research; L.H. performed experiments; M.J. validated results; S.Z. developed software; J.L. analyzed data; Z.H. reviewed and edited manuscript.}
\authordeclaration{The authors declare no conflict of interest.}
% \equalauthors{\textsuperscript{1}A.O.(Author One) contributed equally to this work with A.T. (Author Two) (remove if not applicable).}
\correspondingauthor{\textsuperscript{1}To whom correspondence should be addressed. E-mail: zyhe@dlut.edu.cn}

% % At least three keywords are required at submission. Please provide three to five keywords, separated by the pipe symbol.

\begin{abstract}
Cluster analysis is a fundamental research issue in statistics and machine learning. In many modern clustering methods, we need to determine whether two subsets of samples come from the same cluster. Since these subsets are usually generated by certain clustering procedures, the deployment of classic two-sample tests in this context would yield extremely smaller $p$-values, leading to inflated Type-I error rate. To overcome this bias, we formally introduce the two-cluster test issue and argue that it is a totally different significance testing issue from conventional two-sample test. Meanwhile, we present a new method based on the boundary points between two subsets to derive an analytical $p$-value for the purpose of significance quantification. Experiments on both synthetic and real data sets show that the proposed test is able to significantly reduce the Type-I error rate, in comparison with several classic two-sample testing methods. More importantly, the practical usage of such a two-cluster test is further verified through its applications in tree-based interpretable clustering and significance-based hierarchical clustering.
\end{abstract}

% \dates{This manuscript was compiled on \today}
% \doi{\url{www.pnas.org/cgi/doi/10.1073/pnas.XXXXXXXXXX}}

\maketitle
\ifthenelse{\boolean{shortarticle}}{\ifthenelse{\boolean{singlecolumn}}{\abscontentformatted}{\abscontent}}{}

\firstpage{7}
% Use \firstpage to indicate which paragraph and line will start the second page and subsequent formatting. In this example, there are a total of 11 paragraphs on the first page, counting the first level heading as a paragraph. The value {12} represents the number of the paragraph starting the second page. If a paragraph runs over onto the second page, include a bracket with the paragraph line number starting the second page, followed by the paragraph number in curly brackets, e.g. "\firstpage[4]{11}".

% If your first paragraph (i.e. with the \dropcap) contains a list environment (quote, quotation, theorem, definition, enumerate, itemize...), the line after the list may have some extra indentation. If this is the case, add \parshape=0 to the end of the list environment.
\dropcap{I}n this paper, we formally introduce a new hypothesis testing problem in the context of cluster analysis, namely two-cluster test. Given two disjoint subsets of samples from a data set that is composed of one or more clusters, the objective of such a test is to determine whether they belong to the same cluster or not.

Beyond its theoretical interest, the two-cluster test is a central task in significance-based hierarchical clustering \cite{kimes2017statistical,grabski2023significance,sant2025choir} and tree-based interpretable clustering \cite{hu2025significance,he2025significance,hu2025interpretable}. In these methods, such a test must be conducted to avoid the possibility of over- or under-clustering during the clustering tree growth process \cite{grabski2023significance,sant2025choir}.

\begin{figure}[t]           % “!t” 强制尽可能放在本页页顶
\centering
\includegraphics[width=0.7\linewidth]{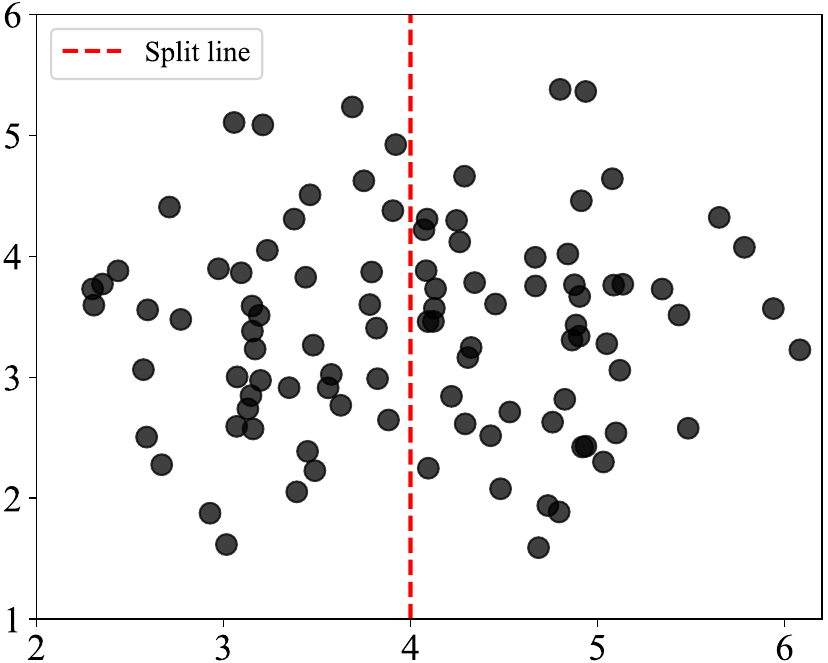}
\caption{Data randomly generated from a Gaussian distribution is split into two subsets by the red line. Obviously, these two subsets of samples belong to the same cluster. The standard Friedman–Rafsky two-sample test returns a $p$-value of 0, indicating that it wrongly rejects the null hypothesis that they originate from the same cluster. In contrast, our two-cluster test method yields a $p$-value of 0.853. }
\label{fig:1}
\end{figure}

One may argue that we can apply existing two-sample test methods \cite{hollander2013nonparametric} to solve the above issue. However, two-sample tests control the Type-I error rate with respect to the mean difference when two subsets are defined a priori \cite{chen2023selective,gao2024selective}. In the above applications, the subsets of samples are instead generated by clustering methods. Therefore, directly applying a classical two-sample test is not feasible, which can yield an extremely inflated Type-I error rate, as illustrated in Fig.~\ref{fig:1} and observed in \cite{hu2025interpretable,hu2025significance}. In order to apply the two-sample test to obtain a valid inference result in this context, one may seek to employ the selective inference technique \cite{chen2023selective,gao2024selective}. However, the selective inference method is highly dependent on the clustering algorithm employed (e.g. $k$-means \cite{chen2023selective} and hierarchical clustering \cite{gao2024selective}), making it difficult to become a general significance test. 

In significance-based hierarchical clustering \cite{kimes2017statistical,grabski2023significance,sant2025choir}, such two-cluster test issue is properly tackled via a simulation-based approach. That is, Monte Carlo sampling is used to generate simulated data sets whose cluster assignments are compared to those of the original data to compute empirical $p$-values. Although intuitive, this procedure incurs substantial computational cost and it cannot provide an analytical $p$-value.

Ideally, we would like enable an analytical $p$-value for any two subsets of samples under the assumption that they belong to the same cluster. However, there are still no research efforts towards this direction and such a two-cluster test issue is rarely known in the literature.

To fill this gap, we formally formulate the two-cluster test issue and present a new significance testing procedure. The new significance testing method is called BTCT (Boundary-based Two-Cluster Test). Specifically, BTCT defines boundary points as samples that are mutual $k$-nearest neighbors between the two subsets. Under the null hypothesis that the two subsets originate from the same cluster, for each boundary point, the number of its $k$-nearest neighbors that fall into the same subset is used as the test statistic, which follows a Binomial($k$, 0.5) distribution. All boundary point $p$-values are then combined via meta-analysis to obtain a census $p$-value.

We examine the validity and effectiveness of BTCT through experiments on both synthetic and real data sets. More importantly, it is applied to significance-based hierarchical clustering and tree-based interpretable clustering, demonstrating its potential in these applications. 
\section*{Problem Statements}
\subsection*{Notation}
Some notations used throughout this paper are summarized in Table \ref{tab:1}. Suppose the given data set $X=\{x_i\}_{i=1}^n\subset\mathbb{R}^d$ is composed
of $K$ clusters. Let $A = \{x_i\}_{i=1}^{n_A}$ and $\ B = \{x_j\}_{j=1}^{n_B}$ be two disjoint subsets in $X$, such that $A \cap B = \emptyset$. For notational simplicity, we denote by $\pi(\cdot)\in\{1,2,...,K\}$ the cluster identifier of any subset under the assumption that all samples in the given subset belong to the same cluster.
\subsection*{Problem Formulation} Given two disjoint subsets $A$ and $B$ from a data set composed of $K$ clusters, our objective is to determine whether these two subsets belong to the same cluster. Specifically, we have the following null and alternative hypotheses:
\begin{equation}
\begin{aligned}
H_0:&\;\pi(A) = \pi(B),\\
H_1:&\;\pi(A) \neq \pi(B).
\label{eqa:1}
\end{aligned}
\end{equation}

\begin{table}[h]
\centering
\caption{Summary of notations.}
\setlength{\tabcolsep}{1.8pt}
\begin{tabular}{ll}
\toprule
  Notation & Definition \\
    \midrule
    $X=\{x_i\}_{i=1}^n\subset\mathbb{R}^D$ & Data set\\
    $A = \{x_i\}_{i=1}^{n_A},\ B = \{y_j\}_{j=1}^{n_B}$ & Two disjoint subsets of $X$  \\
    $K$ & Number of clusters\\
    $n$ & Number of samples\\
    $n_A,n_B$ & Sample sizes of subsets $A$ and $B$, respectively \\
    $\pi(\cdot)$ & The cluster identifier of a subset of samples\\
    $Z =\{z_i\}_{i=1}^{m} = A \cup B$ & The pooled sample set by merging $A$ and $B$ \\
    $m = n_A + n_B$ & Number of samples in $Z$ \\
    $k$ & Number of nearest neighbors \\
    $\mathcal{N}_k(z_i)$ & The set of the $k$ nearest neighbors of $z_i$ in $Z$ \\
    $\mathcal{B}$    & The set of boundary points\\
    $b$              & Number of boundary points \\
    \bottomrule
    \label{tab:1}
  \end{tabular}
\end{table}
\section*{Method}
\subsection*{Boundary point}
To address the hypothesis testing problem formulated in Eq. \ref{eqa:1}, we begin by introducing the notion of boundary points. Specifically, let $Z =\{z_i\}_{i=1}^{m}$ be the pooled sample set by merging \( A = \{x_i\}_{i=1}^{n_A} \) and \( B = \{y_j\}_{j=1}^{n_B} \), where \( m = n_A + n_B \). We define the subset label of a sample $z_{i}$ in $Z$ as:
\begin{equation}
\ell(z_i) =
\begin{cases}
1, & \text{if } z_i \in A, \\
2, & \text{if } z_i \in B.
\end{cases}
\label{eqa:2}
\end{equation}
For each sample $z_i\in Z$, $\mathcal{N}_k(z_i)$ represents the set of its $k$ nearest neighbors in $Z$.

Two samples $z_i$ and $z_j$ are said to be mutual $k$-nearest neighbors if and only if each lies within the other’s set of $k$-nearest neighbors. That is, $z_i \in \mathcal{N}_k(z_j)$ and $z_j \in \mathcal{N}_k(z_i)$. In particular, when $k =1$, we define the boundary point set $\mathcal{B}\subset Z$ as the set of samples in which each sample and its mutual nearest neighbor have different subset labels:
% \begin{equation}
% \begin{aligned}
% \mathcal{B} = \left\{ z_i \in Z \;\middle|\; \exists\, z_j \in Z,\; \mathcal{N}_1(z_i) = z_j , \mathcal{N}_1(z_j) = z_i ,\ell(z_i)\neq \ell(z_j)\right\}.
% \label{eqa:3}
% \end{aligned}
\begin{align}
\mathcal{B} = \{\,z_i \in Z \mid{}&\exists\,z_j \in Z,\;\mathcal{N}_1(z_i)=z_j,\;\mathcal{N}_1(z_j)=z_i,\nonumber\\
&\ell(z_i)\neq \ell(z_j)\,\}.
\end{align}
% \end{equation}

\subsection*{Test statistics and $p$-value calculation} 
% For each boundary point $z_i \in \mathcal{B}$, we use an indicator function $I(\cdot)$ to identify whether the sample and its mutual $k$-nearest neighbor originate from the same source subset.
% \begin{equation}
% I_i(k) =
% \begin{cases}
% 1, & \text{if } z_i \text{ and } \mathcal{N}_k(z_i) \text{ belong to the same subset}, \\
% 0, & \text{otherwise}.
% \end{cases}
% \label{eqa:3}
% \end{equation}
For each boundary point \(z_i \in \mathcal{B}\), the test statistic is defined as follows:
\begin{equation}
T_{i,k} = \sum_{z_{j} \in \mathcal{N}_{k}(z_{i})}
I\bigl(\ell(z_{i}) = \ell(z_{j})\bigr),
\label{eqa:5}
\end{equation}
where $I(\cdot)$ denotes the indicator function. $T_{i,k}$ in Eq. \ref{eqa:5} represents the number of samples in the set of $k$ nearest neighbors of $z_{i}$ whose subset labels are identical to that of $z_{i}$.

Under the null hypothesis that $\pi(A) = \pi(B)$, each neighbor of a boundary point has an equal probability of falling into either subset. Therefore, the test statistic \(T_{i,k}\) follows a binomial distribution:
\begin{equation}
T_{i,k} \sim \mathrm{Bin}(k,0.5).
\label{eqa:6}
\end{equation}

Consequently, the probability of observing at least $t$
neighbors from the same subset can be used as the $p$-value:
\begin{equation}
p_i = \mathbb{P}(T_{i,k} \geq t) = \sum_{r = t}^{k} \binom{k}{r} \left( \frac{1}{2} \right)^k.
\label{eqa:7}
\end{equation}
\subsection*{$P$-value combination}
To assess whether the two subsets belong to the same cluster, we combine the \(p\)-values computed at individual boundary points by applying Fisher’s method \cite{hogg2013introduction}. The combined test statistic is defined as:
\begin{equation}
T_{\text{Fisher}} = -2 \sum_{i=1}^{b} \log(p_i),
\label{eq:fisher}
\end{equation}
where \( b \) is the total number of boundary points.

Under the null hypothesis that all \( p_i \) are independently and uniformly distributed over \([0,1]\), the combined statistic follows a chi-squared distribution with \( 2b \) degrees of freedom:
\begin{equation}
T_{\text{Fisher}} \sim \chi^2(2b).
\label{eq:chisq}
\end{equation}
\begin{table}[h]
	\centering
	\caption{The main characteristics of 23 data sets.}
	\setlength{\tabcolsep}{4mm}
	\begin{tabular}{lllll}
		\toprule
		\label{table2}
		\textbf{Dataset} & $\boldsymbol{n}$ & $\boldsymbol{d}$& $\boldsymbol{K}$ &\textbf{Sources} \\ 
		\midrule
            Zelink6  & 238 & 2&3 &\cite{handl2006multi}\\
            Flame  & 240 & 2&2 &\cite{fu2007flame}\\          
            Zelink1  & 299 & 2&3&\cite{handl2006multi} \\
		Pathbased  & 300 & 2&3 &\cite{chang2008robust}\\
		Jain  & 373& 2&2 &\cite{jain2005data}\\
            Compound  &399 &2 &6 &\cite{zahn1971graph}\\
            Isun  & 400 & 2&3 &\cite{ultsch2005clustering}\\
            Zelink5 & 512 & 2& 4&\cite{handl2006multi} \\
		R15 & 600 & 2&15 &\cite{veenman2002maximum}\\ 
            Aggregation  &788 &2 &7 & \cite{gionis2007clustering}\\
            D31 &3100 &2 &31& \cite{veenman2002maximum}\\
            \midrule
		Iris &150 & 4& 3 &\cite{dua2017uci}\\ 
		Wine&178 &13 &3 &\cite{dua2017uci}\\
            Sonar &208&60&2&\cite{dua2017uci}\\
		Seeds & 210&7 &3 &\cite{dua2017uci}\\
		Glass Identification &214&9&6&\cite{dua2017uci}\\
		New-thyroid & 215& 5& 3 &\cite{dua2017uci}\\
		Ecoli & 336&8 &8  &\cite{dua2017uci}\\
            Ionosphere &351& 33&2 &\cite{dua2017uci}\\
		User Knowledge &403&5&4&\cite{dua2017uci}\\
            HCV&1385&28&5&\cite{dua2017uci}\\
            Yeast &1484&8&10&\cite{dua2017uci}\\
            Segmentation&2310&19&7&\cite{dua2017uci}\\
		\bottomrule
	\end{tabular}
\end{table}
\begin{figure*}[t]           % “!t” 强制尽可能放在本页页顶
\centering
\includegraphics[width=\linewidth]{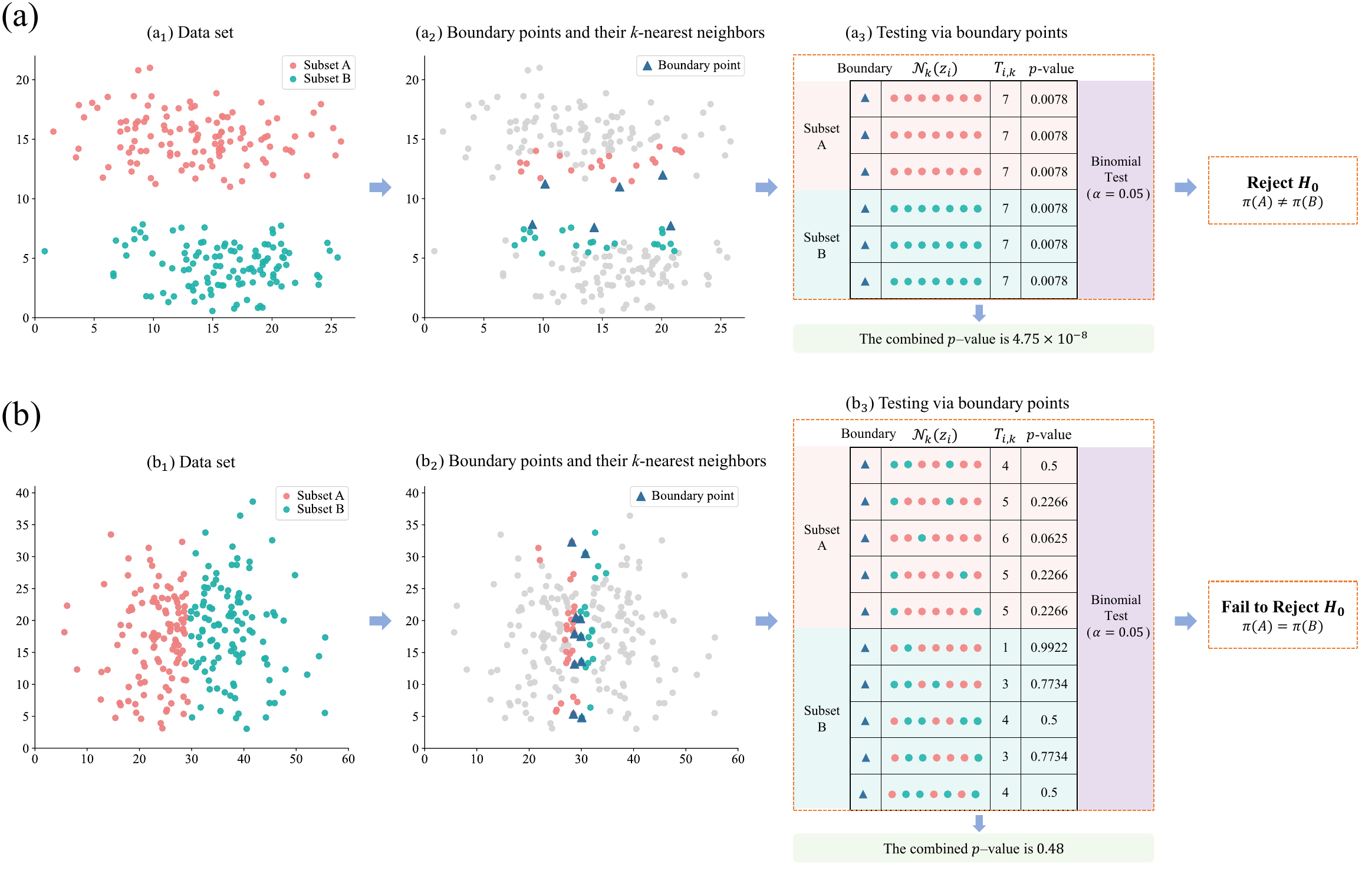}
\caption{An Illustrative Example of BTCT. \textbf{(a)} Example of two subsets from different clusters: (a$_1$) A randomly generated data set with two subsets of 120 samples each. (a$_2$) Identification of boundary points and their $k=7$ nearest neighbors. (a$_3$) For each boundary point, we compute the test statistic $T_{i,k}$ by counting how many of its neighbors belong to the same subset, and obtain the corresponding $p$-value under $\mathrm{Bin}(7,0.5)$. These $p$-values are then combined using Fisher’s method, yielding a combined $p$-value$=4.75\times 10^{-8} < \alpha$, which leads to the rejection of the null hypothesis. \textbf{(b)} Example of two subsets from the same cluster: (b$_1$) A randomly generated data set with 240 samples, which is split into two equally sized subsets along the median of the $x$-axis. (b$_2$) Boundary points and their $k=7$ nearest neighbors are identified. (b$_3$) The combined $p$-value$=0.48 > \alpha$ is computed as before, leading us to fail to reject $H_0$.}
\label{fig:2}
\end{figure*}
\subsection*{An illustrative example of BTCT}
To further illustrate BTCT, we present an example in Fig.~\ref{fig:2}, considering two cases: (a) two subsets from different clusters, and (b) two subsets from the same cluster. 

In Fig.~\ref{fig:2}(a), we first identify boundary points between the two subsets, and then find their $k=7$ nearest neighbors, as shown in Fig.~\ref{fig:2}(a$_2$). For each boundary point, we count how many neighbors belong to the same subset to calculate the test statistic. Under the null hypothesis that the subsets $A$ and $B$ are from the same cluster, the statistic $T_{i,k}$ follows a $\mathrm{Bin}(7,0.5)$ distribution. As shown in Fig.~\ref{fig:2}(a$_3$), all the $7$ nearest neighbors of each boundary point come from the same subset, yielding small $p$-values of 0.0078. The combined $p$-value is $4.75\times 10^{-8}$ by using Fisher’s method, which is less than the significance threshold $\alpha=0.05$, leading us to reject the null hypothesis and indicating that the two subsets do not belong to the same cluster.

Similarly, in Fig.~\ref{fig:2}(b), we repeat the same procedure as before. From Fig.~\ref{fig:2}(b$_3$), we can observe that the $7$ nearest neighbors of most boundary points are equally distributed among two subsets, leading to larger $p$-values. The combined $p$-value$=0.48 > \alpha$, so we fail to reject the null hypothesis, indicating that the two subsets belong to the same cluster.

\section*{Result}
\subsection*{Data sets}
We perform a comprehensive evaluation of our proposed test on 23 data sets. The key characteristics of these data sets are summarized in Table \ref{table2}, sorted in a non-decreasing order of the number of samples. The first 11 data sets are synthetic ones, while the remaining 12 real-world data sets are obtained from the UCI Machine Learning Repository \cite{dua2017uci}.

\subsection*{Baseline methods}
To validate our method, we compare it against three classic two-sample tests, all implemented using the Torch Two Sample package \footnote{\url{https://github.com/josipd/torch-two-sample}}: FR \cite{friedman1979multivariate}, MMD \cite{gretton2012kernel} and Energy \cite{szekely2013energy}. For these methods, we compute the $p$-values using permutation testing with 1000 permutations.

The FR test constructs a minimum spanning tree (MST) over the pooled samples and counts the number of edges that connect samples from different groups, thereby testing whether the two sets of samples originate from distinct distributions.

The MMD test measures the distance between two distributions by comparing their mean embeddings in a reproducing kernel Hilbert space (RKHS).

The Energy test uses the energy distance to assess whether two sets of samples originate from different distributions.

\subsection*{Performance comparison}
This section presents a performance comparison of BTCT and baseline methods on 23 data sets, including both synthetic and real-world data. We examine two main aspects: two subsets from the same cluster and two subsets from different clusters. In either case, we operate under the assumption that all samples within any given subset originate from the same cluster.

\begin{figure}[!t]           % “!t” 强制尽可能放在本页页顶
\centering
\includegraphics[width=\linewidth]{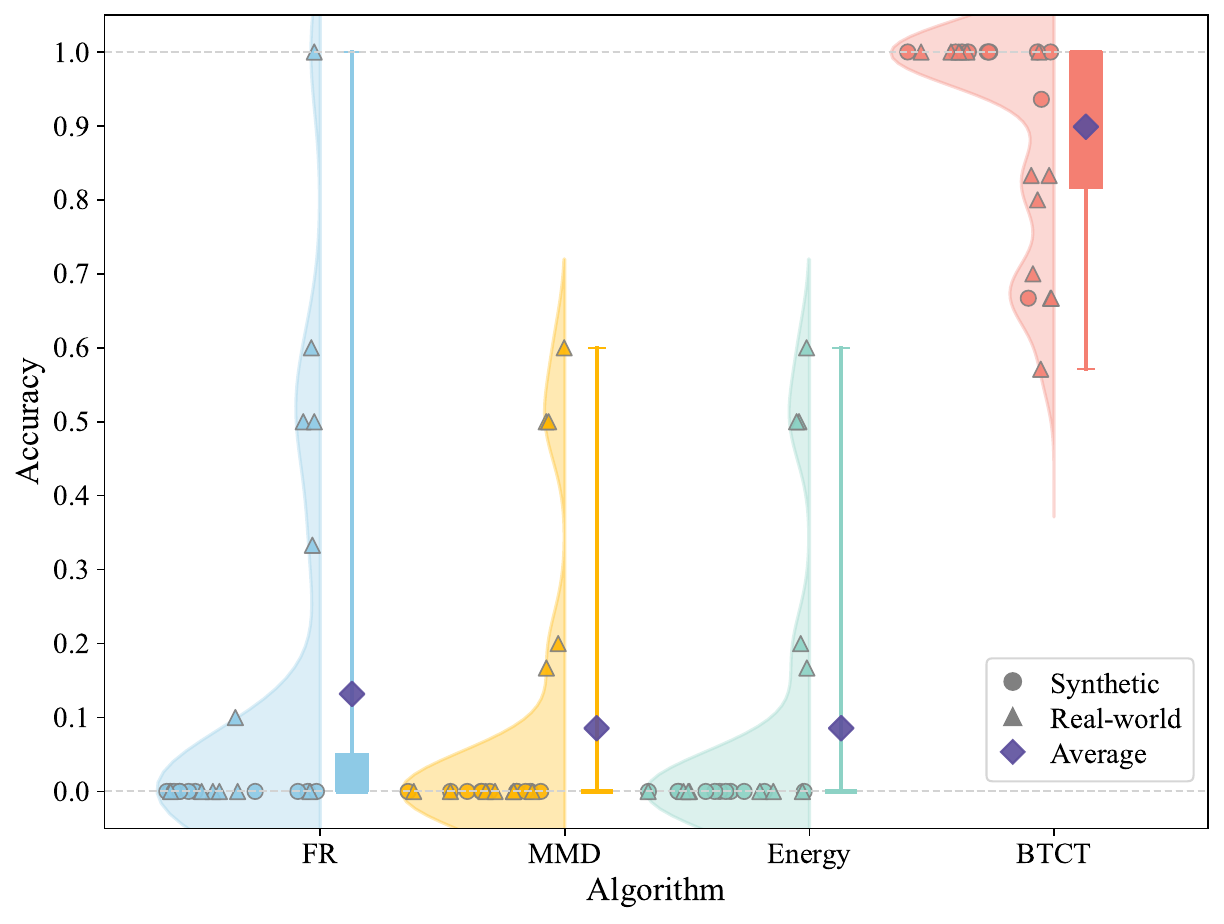}
\caption{Performance comparison of four methods on subset pairs from the same cluster in terms of accuracy. For each algorithm, the left side displays a violin plot with individual data points representing the accuracy on each data set (circles for synthetic data sets and triangles for real-world data sets). On the right, box plots illustrate the interquartile range, min and max values, with diamonds showing the average accuracy.}
\label{fig:3}
\end{figure}
\subsubsection*{Subsets from the same cluster}
For each data set, we apply min-max normalization to ensure consistency in scale, then split each individual cluster into two equally sized subsets based on the median of the first feature values and test whether these two subsets are still identified as belonging to the same cluster. For the MMD test, we use a Gaussian kernel with its bandwidth parameter being equal to the median of pairwise euclidean distances. In BTCT, we set the number of neighbors for each boundary point to $k = 7$. The accuracy is calculated as the ratio of subset pairs that are correctly identified based on their $p$-values. Specifically, a subset pair is correctly identified if the corresponding $p$-value is less than (or larger than) the significance threshold of 0.05 and they come from the different clusters (or the same cluster). Detailed results are presented in Fig.~\ref{fig:3}.

As shown in Fig.~\ref{fig:3}, it is evident that for most data sets, the classic two-sample tests yield accuracies that are zeros, 
with an overall average accuracy of only around 0.1. This is mainly because these methods produce extremely small $p$-values even when the two subsets come from the same cluster, resulting in a highly inflated Type-I error rate. In contrast, BTCT achieves perfect accuracy on most data sets, with an average accuracy of 0.899, demonstrating its ability to correctly recognize subsets from the same cluster.

\begin{figure}[t]           % “!t” 强制尽可能放在本页页顶
\centering
\includegraphics[width=\linewidth]{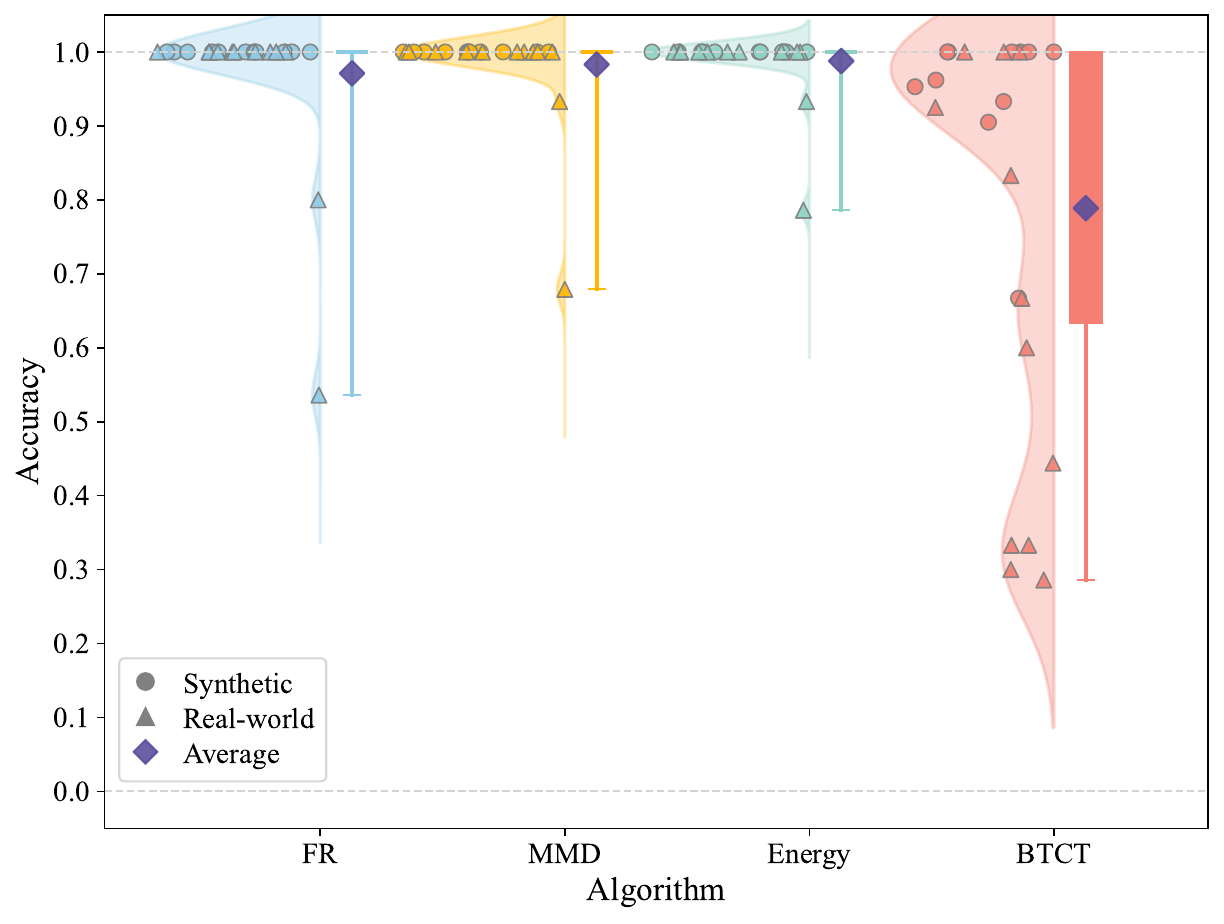}
\caption{Performance comparison of four methods on subset pairs in which each subset corresponds to a ground-truth cluster. The violin plots and box plots can be interpreted in the same manner as that of Fig.~\ref{fig:3}. }
\label{fig:4}
\end{figure}

\subsubsection*{Subsets from different clusters}
We also conduct experiments on all possible pairs of clusters within each data set. As shown in Fig.~\ref{fig:4}, for synthetic data, all two-sample tests achieve an accuracy of 1. Similar to the previous case, the computed $p$-values are extremely small, leading to a rejection of the null hypothesis. BTCT also performs well on synthetic data, indicating that it has high statistical power as well. 

For real-world data, the baseline methods similarly maintain high accuracies close to 1 on most data sets at the cost of failing to distinguish subsets from the same cluster. In contrast, BTCT achieves an average accuracy of 0.789. This is mainly because BTCT generally produces larger $p$-values than classic two-sample tests, a tendency that becomes more pronounced in high-dimensional data.

\section*{Applications}
The two-cluster test can be further applied to relevant tasks such as tree-based interpretable clustering and significance-based hierarchical clustering, in which the classic two-sample test is inappropriate.
\subsection*{Tree-based interpretable clustering}
Currently, tree-based interpretable clustering methods \cite{he2025significance,hu2025interpretable,hu2025significance} typically assess the statistical significance of candidate splits using two-sample tests. However, such tests often yield excessively small $p$-values, leading to potentially unreliable splits. Therefore, we apply BTCT instead of a two-sample test in this context to identify truly meaningful splits to guide the tree growth, thereby avoiding the possibility of over- or under-clustering. In the following, we illustrate this application with an example.

\begin{figure*}[!t]           % “!t” 强制尽可能放在本页页顶
\centering
\includegraphics[width=\linewidth]{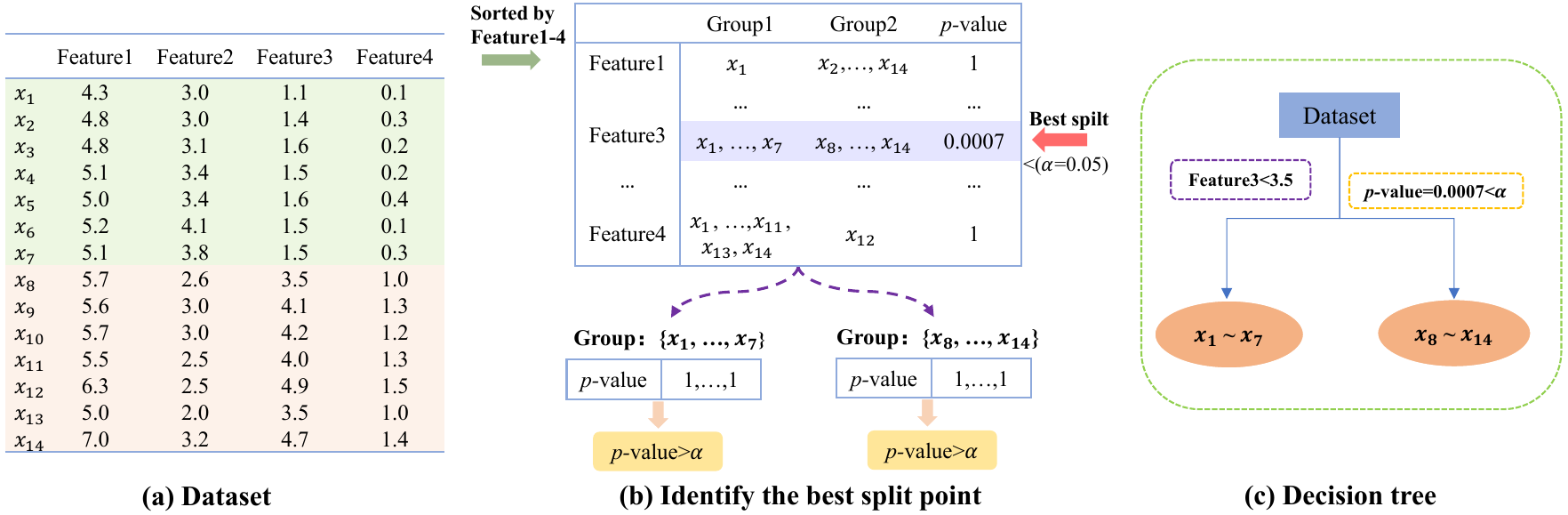}
\caption{Illustration of the tree-based interpretable clustering: (a) A toy data set is constructed by randomly selecting 7 samples from each of two different clusters in the Iris data set. (b) For each feature, samples are sorted in a non-decreasing order. At each candidate split, the data set is divided into two groups and a $p$-value is computed. The best split is selected as the one with the smallest $p$-value among all possible splits. For instance, the best split is found on feature 3 at the root node, which divides 14 samples into two subsets: \{$x_1, \dots, x_7$\} and \{$x_8, \dots, x_{14}$\} with a $p$-value of $0.0007 < \alpha$. Subsequently, the same procedure is recursively applied to the left and right child nodes, in which no splits can yield a $p$-value below $\alpha$. (c) The final clustering tree is composed of two leaf nodes, which correspond to two clusters.}
\label{fig:5}
\end{figure*}
As depicted in Fig.~\ref{fig:5}, for each feature, samples are sorted in a non-decreasing order, and all possible split points that can partition the data into two subsets are examined. For each candidate split, we examine whether the two corresponding subsets belong to the same cluster or not based on either a two-cluster test or a two-sample test. The smallest $p$-value across all candidate splits is then compared to the significance threshold $\alpha = 0.05$: if it falls below this threshold, the corresponding split is considered to be significant to form a branch node; otherwise, the current node is treated as a leaf node. Recursion prioritizes the left child of any branch node, which serves as the new input for this tree growth process, continuing until no leaf nodes can be further divided.

To evaluate the effectiveness of BTCT in tree-based interpretable clustering, we conduct a detailed comparison of BTCT with three classic two-sample tests on 12 real-world data sets. Performance is assessed from two perspectives: clustering quality and interpretability.
\subsubsection*{Performance comparison on clustering quality}
\begin{figure}[t]           % “!t” 强制尽可能放在本页页顶
\centering
\includegraphics[width=\linewidth]{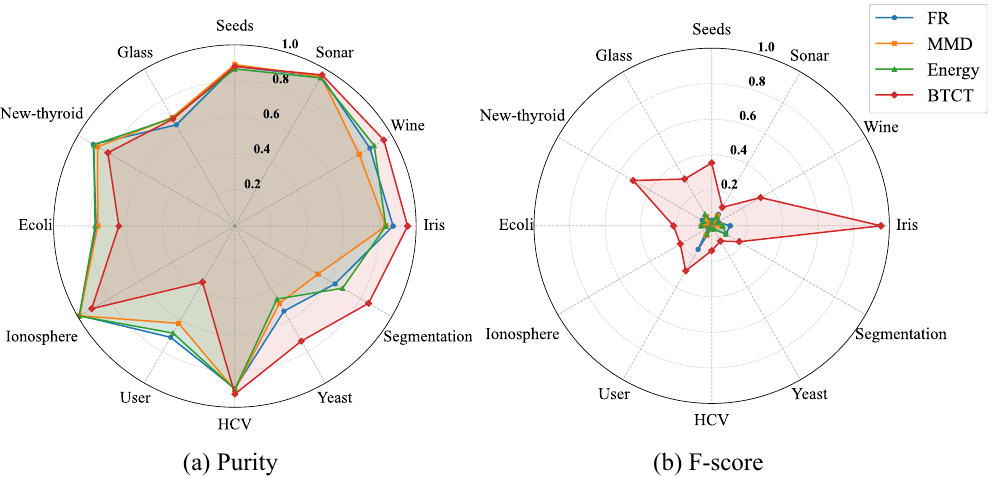}
\caption{Performance comparison of clustering quality across four methods on 12 data sets, evaluated in terms of both purity and F-score.}
\label{fig6}
\end{figure}

Fig.~\ref{fig6} compares the clustering quality of four methods based on Purity and F-score. As shown in Fig.~\ref{fig6}(a), BTCT performs comparably to the three classic two-sample tests across most data sets, with all four methods achieving an average purity of around 0.8. In Fig.~\ref{fig6}(b), BTCT consistently outperforms the other methods on all data sets with respect to the F-score, primarily because baseline methods generate many unnecessary splits due to their excessively small $p$-values. Notably, on the Iris data set, both the purity and F-score achieved by BTCT are close to 1, highlighting the effectiveness of BTCT in tree-based interpretable clustering.

\subsubsection*{Performance comparison on interpretability}
We assess the explainability of clustering results by comparing three metrics: the average depth of the tree (avgDepth), the maximal depth of the tree (maxDepth), and the number of leaf nodes (nLeaf). Simpler tree structures with fewer levels and leaf nodes are generally easier to interpret, which indicates better explainability.

\begin{figure}[h]           % “!t” 强制尽可能放在本页页顶
\centering
\includegraphics[width=\linewidth]{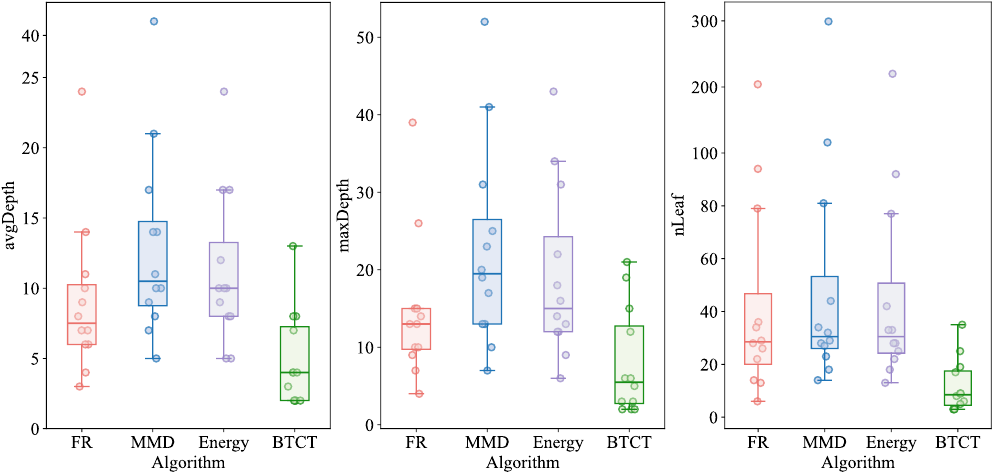}
\caption{Performance comparison of four methods based on three metrics: avgDepth, maxDepth, and nLeaf across 12 data sets.}
\label{fig7}
\end{figure}

As shown in Fig.~\ref{fig7}, BTCT yields an average number of clusters that is closer to the ground-truth number of 5, whereas the other methods significantly overestimate this value. This discrepancy primarily stems from their excessively small $p$-values, which lead to numerous unnecessary splits, ultimately inflating the cluster count. Additionally, BTCT achieves the lowest maxDepth and avgDepth, resulting in shallower trees and thus higher interpretability.

\subsection*{Significance-based hierarchical clustering}
Given a data set $X$ of $n$ samples, hierarchical clustering methods estimate all $K = 1, \dots, n$ partitions of $X$ through a sequential procedure. For instance, the divisive (top-down) approach begins by placing all $n$ samples in one cluster. Then, at each step, one cluster is chosen to be divided into two clusters. After $n-1$ steps, we have $n$ clusters where each cluster is composed of only one sample. 

\begin{figure}[H]           % “!t” 强制尽可能放在本页页顶
\centering
\includegraphics[width=0.9\linewidth]{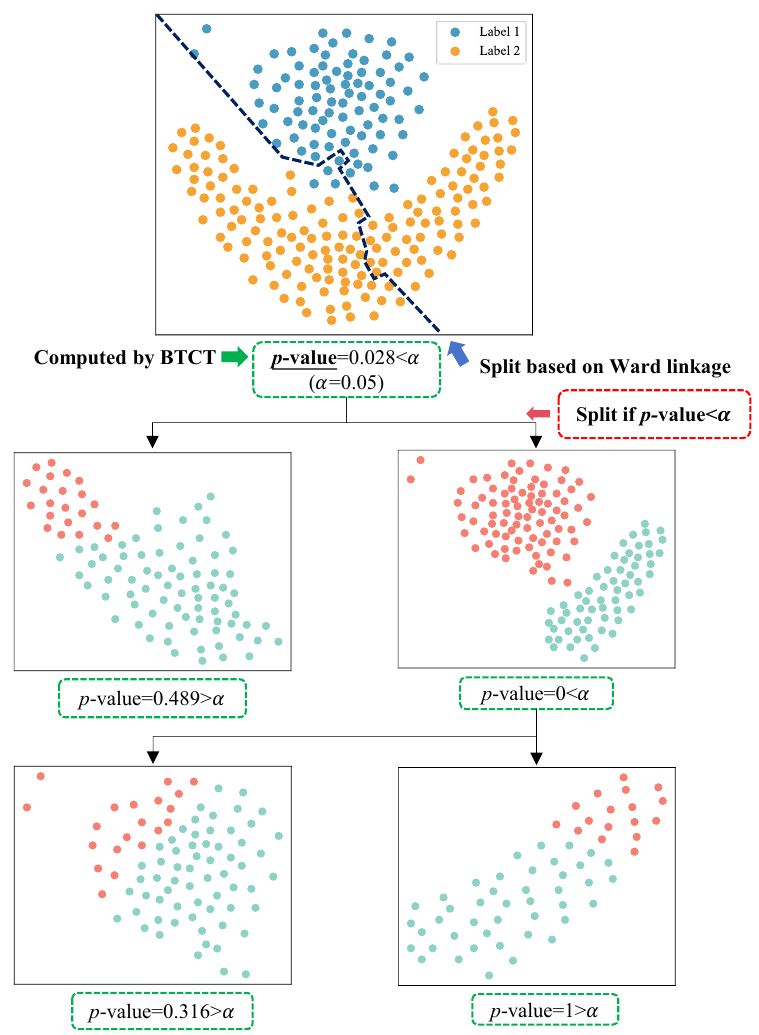}
\caption{Illustration of significance-based hierarchical clustering. The Flame data set is divided into two subsets based on Ward’s criterion at the root node. The corresponding split is considered to be statistically significant since the $p$-value $= 0.028$ is less than $\alpha$. This procedure is recursively applied to each child node, with splits occurring only when the $p$-value is below $\alpha$; otherwise, no further division is made.}
\label{fig8}
\end{figure}

One critical issue in divisive hierarchical clustering is when should we terminate the split operation. If it is terminated earlier or later, then under- or over-clustering will occur. Hence, people begin to employ hypothesis testing methods to assess the statistical significance of each division, thereby yielding a right cluster number estimation \cite{kimes2017statistical,grabski2023significance,sant2025choir}.  The idea is similar to that used in tree-based interpretable clustering, we can assess whether two divided subsets belong to the same cluster via a two-cluster test or a two-sample test. An example is given in Fig.~\ref{fig8}, where we use our method to determine whether the corresponding division is statistically significant.

\begin{figure}[h]           % “!t” 强制尽可能放在本页页顶
\centering
\includegraphics[width=0.8\linewidth]{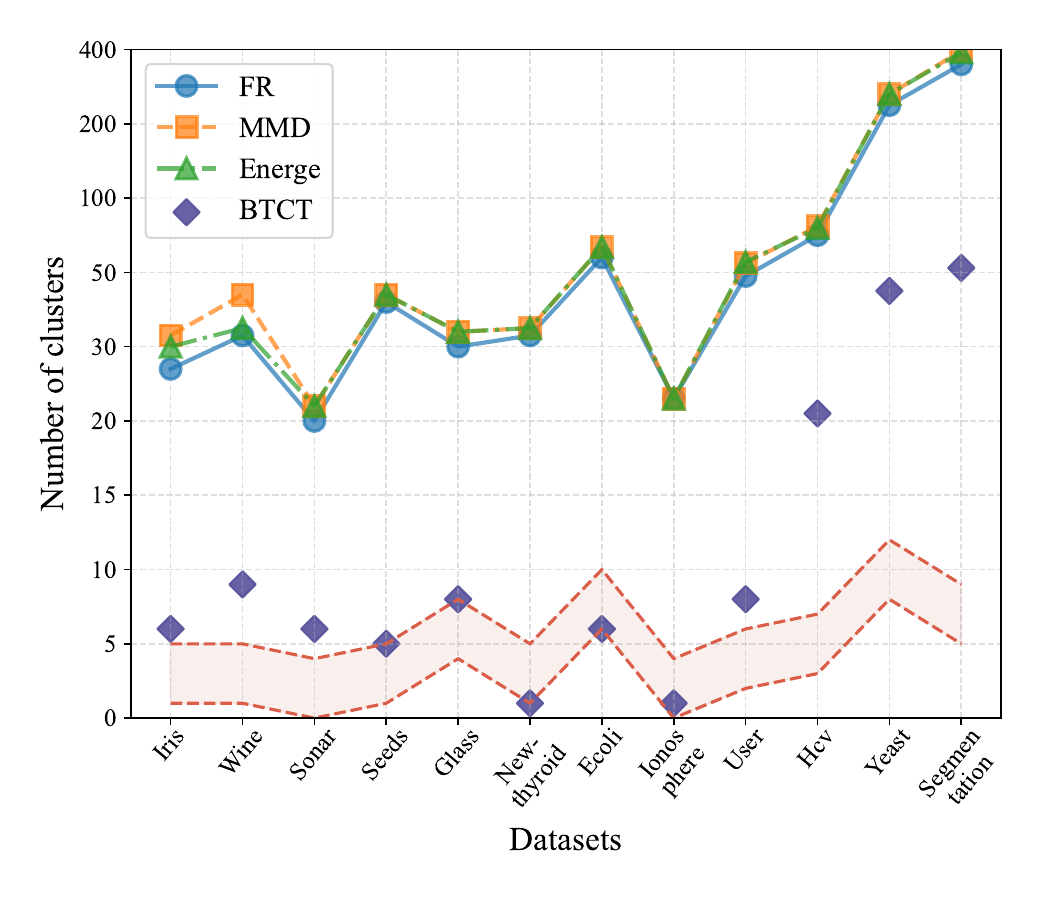}
\caption{Performance comparison of four methods on 12 data sets in terms of predicted cluster number, with the ground-truth cluster number region $K - 2$ to $K + 2$ for each data set highlighted in red.}
\label{fig9}
\end{figure}

\subsubsection*{Prediction of cluster number}
Fig.~\ref{fig9} compares the predicted number of clusters by four methods across 12 data sets. The red shaded region is the interval $[K-2, K+2]$ for each data set, where $K$ is the number of ground-truth clusters.  The predicted cluster number by BTCT is close to this region on most data sets, and slight overestimation occurs due to the impurity of the subsets created by the hierarchical clustering method based on Ward linkage at each split. In contrast, the baseline methods often yield extremely small $p$-values for candidate divisions, leading to over-clustering and overestimation of cluster number.

\section*{Discussion}
We introduce the two-cluster test, a significance testing issue that has been overlooked in the field of cluster analysis. We highlight the importance of such a new test and demonstrate that classic two-sample tests fail to control the Type-I error rate when the objective is to determine whether two subsets of samples come from the same cluster.  More importantly, the two-cluster test is the key issue in modern clustering algorithms such as tree-based interpretable clustering algorithms and significance-based hierarchical clustering algorithms.

There are still many interesting issues that remain unaddressed. Firstly, the performance of the proposed BTCT method is still far from being satisfactory. Hence, more effective methods should be further developed. Secondly, new clustering algorithms that are equipped with two-cluster test should be developed so as to demonstrate its practical usage. Finally, more application scenarios for the two-cluster test should be investigated, such as hierarchical community detection \cite{li2022hierarchical}.  
\section*{Data, Materials, and Software Availability.}
All data sets used in this study are listed in Table \ref{table2} and the source codes of BTCT are available at \url{https://github.com/Xying-Liu/Two-cluster-test}.
% \showmatmethods{} % Display the Materials and Methods section

\acknow{This work has been supported by the Natural Science Foundation of China under Grant No. 62472064.}

\showacknow{} % Display the acknowledgments section

% \bibsplit[2]
%Use \bibsplit to split the references from the body of the text. Value "[2]" represents the number of reference in the left column (Note: Please avoid single column figures & tables on this page.)

% Bibliography
% \bibliography{pnas-sample}
% \bibliographystyle{pnas-new}
% \bibliography{pnas-sample}

\end{document}